\documentclass[sigconf]{acmart}

\usepackage{booktabs} 
\usepackage{multirow}
\usepackage{graphicx}
\usepackage{subfigure}
\usepackage{amsmath}
\usepackage{algorithm}
\usepackage{algorithmic}
\usepackage{makecell}

\setcopyright{rightsretained}

\acmDOI{10.475/123_4}

\acmISBN{123-4567-24-567/08/06}

\acmConference[ICMR19]{ACM International Conference on Multimedia Retrieval}{July 2019}{El
  Paso, Texas USA}
\acmYear{2019}
\copyrightyear{2019}

\acmArticle{4}
\acmPrice{15.00}

\editor{Jennifer B. Sartor}
\editor{Theo D'Hondt}
\editor{Wolfgang De Meuter}

\begin{document}
\title{Feature Pyramid Hashing}

\author{Yifan Yang}
\affiliation{%
 \institution{School of Data and Computer Science\\Sun Yat-Sen University}
 \city{Guangzhou}
 \state{China}
}
\email{yangyf26@mail2.sysu.edu.cn}

\author{Libing Geng}
\affiliation{%
 \institution{School of Data and Computer Science\\Sun Yat-Sen University}
 \city{Guangzhou}
 \state{China}
}
\email{genglb@mail2.sysu.edu.cn}

\author{Hanjiang Lai}
\affiliation{%
 \institution{School of Data and Computer Science\\Sun Yat-Sen University}
 \city{Guangzhou}
 \state{China}
}
\email{laihanj3@mail.sysu.edu}

\author{Yan Pan}
\affiliation{%
  \institution{School of Data and Computer Science\\Sun Yat-Sen University}
  \city{Guangzhou}
 \state{China}
}
\email{panyan5@mail.sysu.edu}

\author{Jian Yin}
\affiliation{%
 \institution{School of Data and Computer Science\\Sun Yat-Sen University}
 \city{Guangzhou}
 \state{China}
}
\email{issjyin@mail.sysu.edu.cn}

\renewcommand{\shortauthors}{Y. Yang et al.}

\begin{abstract}
In recent years, deep-networks-based hashing has become a leading approach for large-scale image retrieval. Most deep hashing approaches use the high layer to extract the powerful semantic representations. However, these methods have limited ability for fine-grained image retrieval because the semantic features extracted from the high layer are difficult in capturing the subtle differences. To this end, we propose a novel two-pyramid hashing architecture to learn both the semantic information and the subtle appearance details for fine-grained image search. Inspired by the feature pyramids of convolutional neural network, a \textit{vertical pyramid} is proposed to capture the high-layer features and a \textit{horizontal pyramid} combines multiple low-layer features with structural information to capture the subtle differences. To fuse the low-level features, a novel combination strategy, called consensus fusion, is proposed to capture all subtle information from several low-layers for finer retrieval. Extensive evaluation on two fine-grained datasets CUB-200-2011 and Stanford Dogs demonstrate that the proposed method achieves significant performance compared with the state-of-art baselines.
\end{abstract}

%


\maketitle

\section{Introduction}
Due to the rapid development of the internet, the amount of images grows rapidly. Image retrieval has attracted increasing interest, especially for the large-scale image databases with millions to billions of images. Hashing methods, which encode data into binary codes, have been widely studied  due to the retrieval efficiency in both storage and computation. In this paper, we focus on deep hashing for fine-grained image retrieval.

Much effort has been devoted to deep-networks-based hashing for large-scale image retrieval  (e.g., ~\cite{wang2018survey,cao2018deep}). These approaches use deep networks to learn similarity-preserving hash functions, and the similar images will be encoded to nearby hash codes. Xia et al.~\cite{xia2014supervised} firstly present a two-stage method for learning good image representation and hash functions. Further, Lai et al.~\cite{lai2015simultaneous} and Zhuang et al.~\cite{zhuang2016fast} proposed to use the triplet ranking loss to preserve the similarities. The deep pairwise methods also proposed to learn the hash functions, e.g., DPSH~\cite{li2015feature} and DSH~ \cite{liu2016deep}. Recently, the generative adversarial networks have been achieved much attention for image retrieval, e.g.,~\cite{lin2018adversarial,zhang2018attention}.

\begin{figure}[t]
\centering
\includegraphics[width=2.8in]{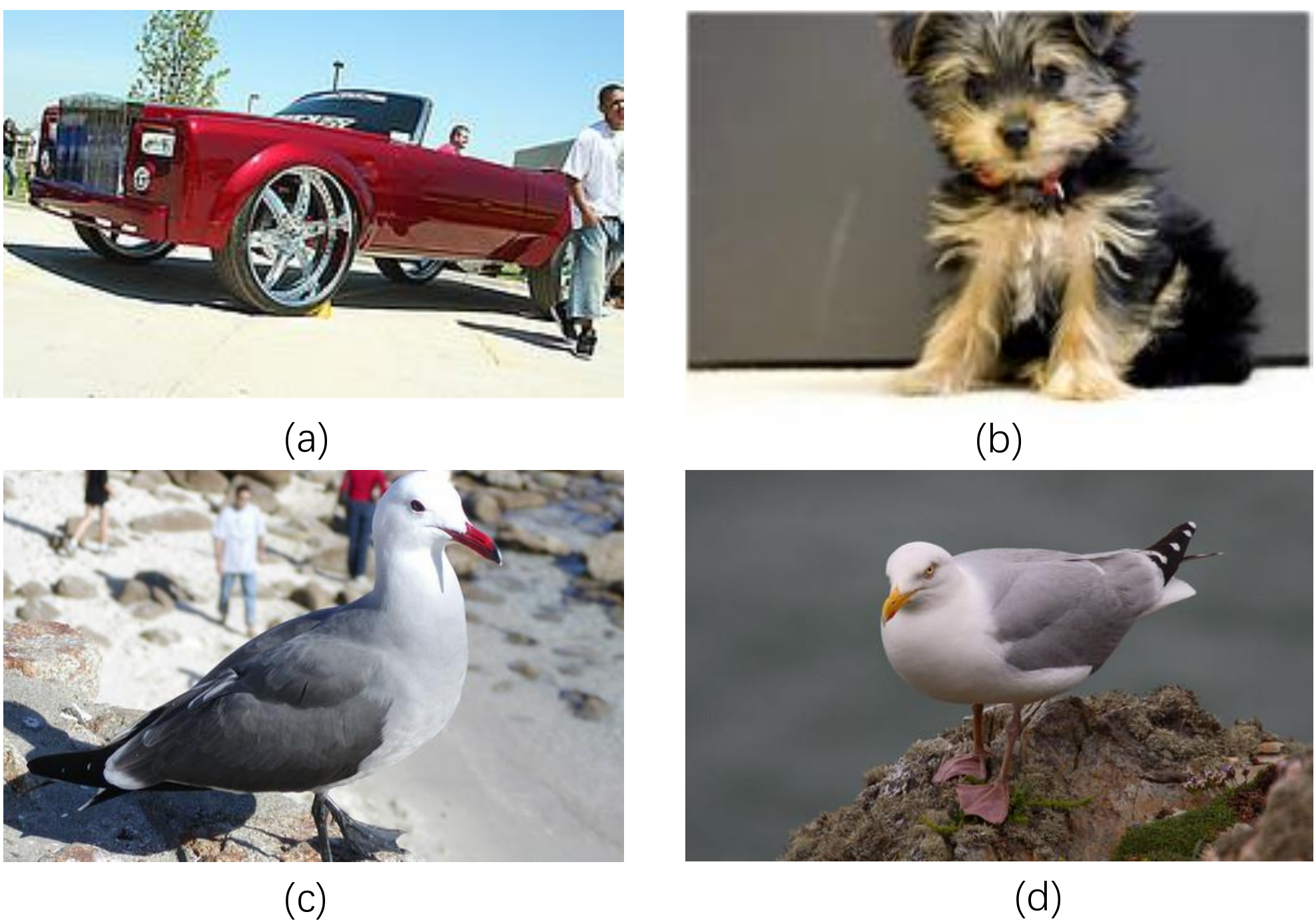}
\caption{(a) and (b) are two images from the coarse-grained dataset; (c) and (d) are images from different classes in CUB-200-2011.}
\label{grainedimage}
\end{figure}

However, most of the existing deep hashing methods are designed for the coarse-grained datasets, e.g., CIFAR-10 and NUS-WIDE. For coarse-grained datasets, the most important thing is to find semantic differences between the images from different categories as shown in Figure~\ref{grainedimage}. Since the high layers of the CNN tend to extract the semantic information~\cite{yu2018hierarchical}, most deep hashing methods utilize the highest layer, e.g., fc7, to extract the power image representations and show impressive performance on the coarse-grained databases. However, for fine-grained objects, only considering the semantic information may not enough. Taken two images of CUB-200-2011 as an example in Figure~\ref{grainedimage}, it is indistinguishable by only using the high-level features. The differences of the fine-grained objects rely on the subtle appearance details such as a small part of the tail and the beak.  When it turns to deep hashing for fine-grained data, the problem translates into how to capture subtle details and embed them into hash codes. 

Feature pyramids~\cite{lin2017feature,chen2018cascaded}, which improve the performance by using different layers in  convolutional network, become a popular approach. U-Net~\cite{ronneberger2015u} presents a contracting path to associate the low-level features and the symmetric high-level features. Feature Pyramid Network (FPN)~\cite{lin2017feature} uses the inherent multi-scale pyramidal hierarchy of deep convolutional networks for object detection, which obtained outstanding accuracy on small-scale objects. Kong et al.~\cite{kong2018deep} proposed a reconfiguration architecture to combine low-level and high-level features in a non-linear way. Although the success, it is still a problem that has not been studied for hashing: how to encode the feature pyramids into efficient binary codes. Recently, Zhao et al.~\cite{zhao2017spatial} proposed a spatial pyramid deep hashing for large-scale image retrieval. Jin~\cite{jin2018deep} showed an attention mechanism to learn the hashing codes for fine-grained data. However, these methods perform the attention or spatial pyramid pooling on the high layer and do not consider that objects have multiple scales. Hence, how to combine the low-level features and capture the subtle appearance differences into binary codes should be more explored.

\begin{figure*}[!h]
\includegraphics[width=7.0in]{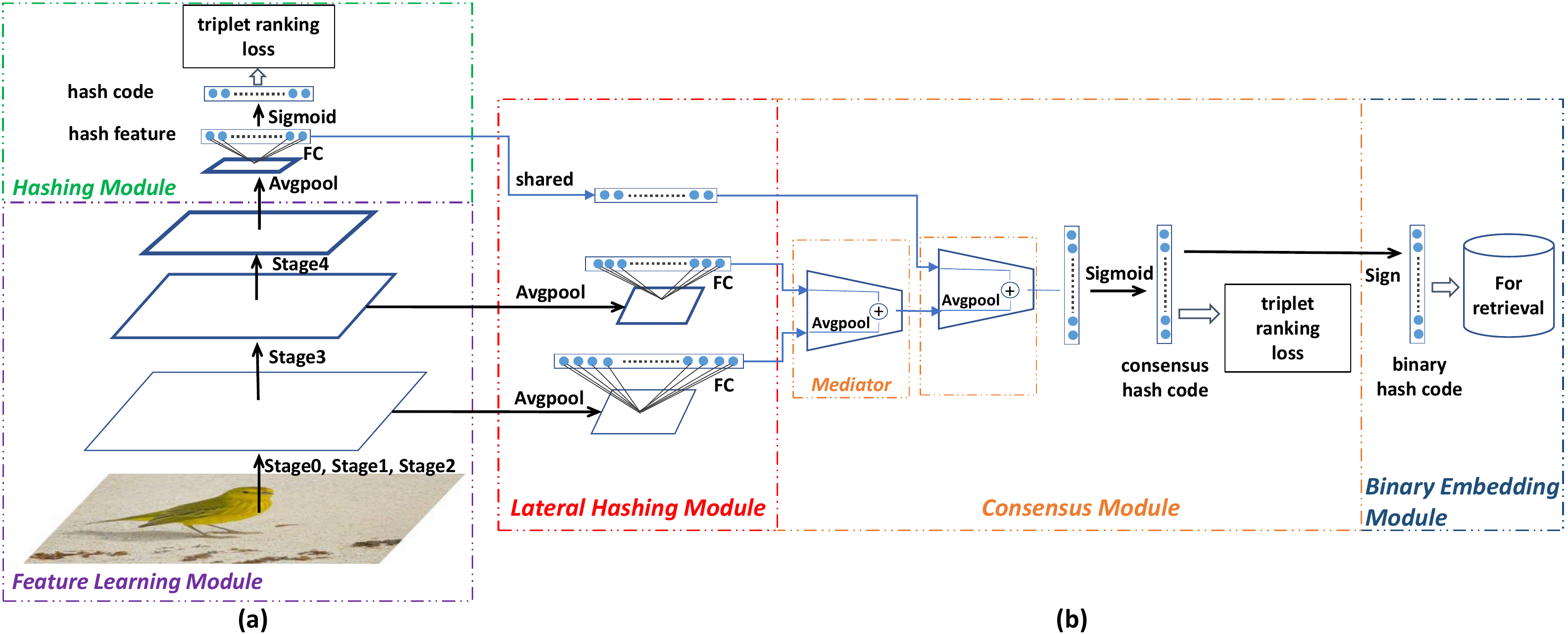}
\caption{Overview of the proposed two-pyramid architecture. (a) The Vertical Pyramid is divided into two parts: the feature learning module and the hashing module. The feature learning module is a convolutional sub-network which maps the input multi-modal data into the high-level feature representations. For the sake of clarity, we divided the convolutional sub-network into several stages, and the details of the division will be presented in Table \ref{resnets_stage}. The hashing module tries to learn powerful image representations. (b)The Horizontal pyramid contains three components: the lateral hashing module, a consensus module and binary embedding module. The lateral hashing module leverage pyramidal feature for hash feature. Then, the consensus module composes a consensus hash feature through the mediator and turns feature to consensus hash code. Finally, the binary embedding module encodes consensus hash codes into binary codes for retrieval. In addition, triplet ranking loss is defined on these hash codes to preserve relative similarities among the images.}
\label{Overview}
\end{figure*}

In this paper, we propose a simple yet efficient two-pyramid architecture using the pyramidal features to compose our hash codes. It highly improves the performance of deep hashing for fine-grained image retrieval. Specially, as shown in  Figure \ref{Overview}, our architecture has two pyramids: the vertical and horizontal pyramids. 1) The vertical pyramid aims to capture the semantic differences between the fine-grained objects. It firstly captures the feature of input images by the sub-network consists of stacked convolution layer, then applying the average pooling layer on the top feature followed by a full connected layer with sigmoid activation to learn the hash code. On the top of the hash code, we use a triplet ranking loss to preserve relative similarities among the input images and maintain the hashing capability throughout the networks holistically. 2) The horizontal pyramid was proposed to capture the subtle details and encode these details into binary codes via consensus learning. As shown in Figure~\ref{Overview} (b), the horizontal pyramid firstly uses the feature maps from different stages of the sub-network to generate hash features by capturing different scales of the objects. A consensus fusion mechanism is proposed to encode all these multi-scale features into one powerful hash code. The consensus fusion mechanism which is composed of two modules in Figure \ref{Overview} (b) includes average pooling layers, fully-connected layers and addition function layers. In the end, we employ the triplet ranking loss to generate the similarity-preserving hash codes for the codes with subtle information.

The main contributions of our work are listed as follows. Firstly, we are one of the first attempts at using ConvNet's pyramidal features hierarchy to compose the hash code for fine-grained retrieval. The proposed method can capture not only the semantic differences but also the subtle differences of fine-grained objects. Secondly, we propose a consensus fusion mechanism to encode all subtle details into the binary codes. Finally, Our architecture obtains significant results on the two corresponding fine-grained datasets in comparison with several state-of-the-art methods.

\section{Related Work}

Hashing method \cite{wang2018survey}, which learns similarity-preserving hash functions to encode data into binary codes, has become a popular approach. Existing methods can be mainly divided into three categories: unsupervised, semi-supervised and supervised methods. 

Unsupervised methods \cite{lin2016learning} attempt to learn similarity-preserving hash functions by utilizing unlabeled data during the training procedure. ITerative Quantization (ITQ) \cite{gong2013iterative}, Anchor Graph Hashing (AGH)  \cite{liu2011hashing}, Kernerlized LSH (KLSH) \cite{kulis2009learning}, Spectral Hashing (SH) \cite{weiss2009spectral} and semantic hashing \cite{salakhutdinov2007learning} are the representative methods. 
Lately, an unsupervised deep hashing approach, named DeepBit, was proposed by Lin et al.~\cite{lin2016learning}. It learns the binary codes by satisfying three criterions on binary codes: minimal quantization loss, evenly distributed codes and uncorrelated bits. Further, a similarity adaptive deep hashing (SADH)~\cite{shen2018unsupervised} was proposed, which alternatively proceeds over three modules: deep hash model training, similarity graph updating and binary code optimization.

Semi-supervised methods make use of the labelled data and the abundant unlabelled data to learn better hashing functions. One of representative work is Semi-Supervised Hashing (SSH) \cite{wang2010semi}, which regularizes the hashing functions over the labelled and the unlabeled data. Sequential Projection Learning for Hashing (SPLH)~\cite{wang2010sequential} is proposed to learn the hash functions in sequence. Xu et al.~\cite{wu2013semi} proposed bootstrap sequential projection learning for nonlinear hashing (Bootstrap-NSPLH).  DSH-GAN~\cite{qiu2017deep} is a deep architecture, which contains a semi-supervised GAN to produce synthetic images, and a deep semantic hashing network with real-synthetic triplets to learn hash functions.

Supervised methods \cite{lin2013general} \cite{lai2015simultaneous} seek to utilize supervised information, e.g., pairwise similarities, or relative similarities of images, to learn better bit wise representations. For example, Minimal loss hashing (MLH) \cite{norouzi2011minimal} uses structural SVMs with latent variables to encodes images. Kernel-based Supervised Hashing (KSH) \cite{liu2012supervised} learns hash functions by minimizing similar pairs' hamming distance and maximized one of the dissimilar pairs. Binary Reconstruction Embedding (BRE) \cite{kulis2009learning} tries to minimizes the reconstruction errors between the Hamming distance of the learned binary codes and the original distances of the data points. The ranking preserving hashing approach~\cite{wang2015ranking} directly optimizes the NDCG measure. 

In recent years, inspired by the significant achievements of deep neural networks, learning the hash codes with deep neural networks (deep hashing) has become a novel stream of supervised hashing methods. For example, 
Lai et al.~\cite{lai2015simultaneous} proposed a deep triplet-based loss function for supervised hashing method. DPSH \cite{li2015feature} is a deep hashing method to perform simultaneous feature learning and hash code learning with pairwise labels. DSH \cite{liu2016deep} speeds up the training of the network by adding a regular term instead of activation function to loss function. HashNet~\cite{cao2017hashnet} utilizes the weighted pairwise loss to maximize the likelihood function and takes a weighted attenuation factor on the activation function. It reduces the semantic loss caused by feature-to-hash code mapping. SPDH-SPBPM~\cite{zhao2017spatial} divides the feature map of the last convolutional layer into several sets of spatial parts. However, these methods are designed for the coarse-grained datasets. Few works~\cite{jin2018deep} have been proposed for the fine-grained image retrieval. Different from these existing fine-grained hashing method which use the high layer features, we combine both the low-level and high-level features into our framework. 




Feature pyramids have achieved great success in many vision tasks. For example, FPN~\cite{lin2017feature} adds the feature maps of the highest-layer to feature maps of several low-layer, and then performs object detection on each layer.  Different from the existing feature pyramid methods, our goal is to generate hash codes. We do not directly generate hash codes from one layer but use multi-level features. A consensus fusion is proposed to combine the multi-level features. 

\section{The Proposed Approach}
We denote $\mathcal{I}$ as the image space. The task of learning-based hashing for images is to learn a mapping function $F: \mathcal{I} \to {\{0,1\}}^q$ such that an input image $I\in\mathcal{I}$ can be mapped to an $q$-bit binary code ${F(I)}$, where the similarities among images are preserved in the Hamming space. 

In this paper, we propose a deep architecture for learning hash codes. As shown in Figure~\ref{Overview}, the proposed architecture has two pyramids from vertical orientation and horizontal orientation, respectively. The vertical pyramid extracts the feature from raw images and maintains the hashing capability throughout the networks holistically. The horizontal pyramid leverages pyramidal features from different stages of CNN for learning the hash feature and then aggregates the hash feature into the final binary code for retrieval. 

\subsection{Vertical Pyramid}
Vertical pyramid contains two components: (1) the feature learning module with stacked convolution layers to capture the effective feature of an input image; (2) the hashing module to maintains the hashing capability throughout the networks holistically. In the following, we will present the details of these components, respectively. 

\subsubsection{\textbf{Feature Learning Module}}
As shown in Figure \ref{Overview}(a), we use a convolutional sub-network with multiple convolution layers as feature learning module to capture a discriminative feature representation of the input images. The feature learning module is based on the architecture of ResNets~\cite{resnet}, which computes a feature hierarchy consisting of feature maps at several scales with a scaling step of 2. There are often many layers producing feature maps of the same size, and we denote these layers are in the same network $stage$. For a clearer description of the network $stage$, Table \ref{resnets_stage} shows the detailed division of the ResNet18 on the $stage$, which follows the same division suggestion of $stage$ in the FPN\cite{lin2017feature}.
\newcommand{\blocka}[2]{\multirow{3}{*}{\(\left[\begin{array}{c}\text{3$\times$3, #1}\\[-.1em] \text{3$\times$3, #1} \end{array}\right]\)$\times$#2}
}
\newcommand{\blockb}[3]{\multirow{3}{*}{\(\left[\begin{array}{c}\text{1$\times$1, #2}\\[-.1em] \text{3$\times$3, #2}\\[-.1em] \text{1$\times$1, #1}\end{array}\right]\)$\times$#3}
}
\renewcommand\arraystretch{1.1}
\setlength{\tabcolsep}{3pt}
\begin{table}
\centering

\caption{$Stage$ of ResNet18}
\label{resnets_stage}
\begin{tabular}{|c|p{2.5cm}<{\centering}|p{1.3cm}<{\centering}|p{3.0cm}<{\centering}|}
\hline

\multirow{1}{*}{{stage}} &\multicolumn{1}{c|}{name in ResNet\cite{resnet}} &\multicolumn{1}{c|}{output size} &\multicolumn{1}{c|}{remarks}\\
\hline

\multirow{1}{*}{{0}} &conv1 &$112\times{112}$ &$7\times{7}$, 64, stride 2 \\\hline

\multirow{4}{*}{{1}} &\multirow{4}{*}{{conv2\_x}} &\multirow{4}{*}{$56\times{56}$} &$3\times{3}$ max pool, stride 2\\
& & &\bf\blocka{64}{2}\\
& & &\\
& & &\\
\hline

\multirow{3}{*}{{2}} &\multirow{3}{*}{{conv3\_x}} &\multirow{3}{*}{$28\times{28}$} &\bf\blocka{128}{2}\\
& & &\\
& & &\\
\hline

\multirow{3}{*}{{3}} &\multirow{3}{*}{{conv4\_x}} &\multirow{3}{*}{$14\times{14}$} &\bf\blocka{256}{2}\\
& & &\\
& & &\\
\hline

\multirow{3}{*}{{4}} &\multirow{3}{*}{{conv5\_x}} &\multirow{3}{*}{$7\times{7}$} &\bf\blocka{512}{2}\\
& & &\\
& & &\\
\hline

\end{tabular}
\end{table}

We define one pyramid level for each stage. The output of the last layer of each stage will serve as the output of each stage and play a role as the side-output in the vertical pyramid. In training, we use the pre-trained ResNet~\cite{resnet} model to initialize the weights in this sub-network. We denote the whole Module as ${E}$ and $\mathbf{m} = E(I)$ for the output of ${E}$ of the input image ${I}$. In order to describe the side outputs feature of the sub-network, let's denote the output feature map of the $[stage1, stage2, stage3, stage4]$ as $[\mathbf{m}^{(s2)} = E^{(s2)}(I), \mathbf{m}^{(s3)} = E^{(s3)}(I),$ $ \mathbf{m}^{(s4)} = E^{(s4)}(I)]$, respectively. Specially, 
${\mathbf{m}^{(s4)} = \mathbf{m}}$. 

\subsubsection{\textbf{Hashing Module}}
Based on our prior knowledge and cross-validation results, the feature maps ${\mathbf{m}^{(s4)}}$ has the highest-level semantics and achieve better performance for retrieval when quantized as a binary hash code compared to other stages' feature. Accordingly, in order to learn powerful image representations, we employed the ${\mathbf{m}^{(s4)}}$ and the hash module in training. Following the traditional deep learning to hash setting, on top of the feature map $\mathbf{m}$, we add a fully connected layer ${FC}$ with sigmoid activation function ${(Sigmoid)}$. Specifically, using the $\mathbf{f}^{(s4)}$ be the output vector of the fully connected layer (i.e., the hash feature), one can obtain the hash code $\mathbf{v}$ by:
\begin{equation} \label{fc1}
  \begin{split}
  \mathbf{a}^{(s4)} &= Avgpool(\mathbf{m}^{(s4)}) ,    \\
  \mathbf{f}^{(s4)} &= FC(\mathbf{a}^{(s4)}) ,
  \end{split}
\end{equation}where $\mathbf{m}^{(s4)}\in{\mathbb{R}^{512\times{7}\times{7}}}$, $\mathbf{a}^{(s4)}\in{\mathbb{R}^{512\times{1}\times{1}}}$ and $\mathbf{f}$ is $q$-dimensional hash feature.
\begin{equation}
\mathbf{v}=Sigmoid(\mathbf{f}^{(s4)}) ,
\end{equation}where ${\mathbf{v}}$ is an $q$-dimensional hash code, each of whose elements is in the range $[0, 1]$, respectively.

For ranking-based image retrieval, it is a common practice to preserve relative similarities of the form ``image ${I_{i}}$ is more similar to image ${I_{j}}$ than to image ${I_{k}}$''. To learn hash codes preserving such relative similarities, the triplet ranking loss has been proposed in the existing hashing methods~\cite{lai2015simultaneous}. Specifically, for a triplet of images $(I_{i},I_{j},I_{k})$ that $I_{i}$ is more similar to $I_{j}$ than to $I_{k}$, we denote the real-valued hash code for $I_{i}$, $I_{j}$ and $I_{k}$ as ${\mathbf{v_i}}$, ${\mathbf{v_j}}$ and ${\mathbf{v_k}}$, respectively. The triplet ranking loss function is defined as:
\begin{equation}\label{loss}
\begin{split}
&\ell_{tri}({\mathbf{v_i}},\mathbf{v_j},\mathbf{v_k})\\
=&\max(0,m_n+||{\mathbf{v_i}}-\mathbf{v_j}||_2^2-||{\mathbf{v_i}}-\mathbf{v_k}||_2^2)\\
&s.t.\ {\mathbf{v_i}},\ \mathbf{v_j},\ \mathbf{v_k}
\in [0,1]^q,\\
\end{split}
\end{equation}where $m_n$ is the margin parameter depending on the hash code length $q$, $||.||_2$ is the $\ell_2$ norm.

Note that the triplet loss in Eq.(\ref{loss}) is designed for single-label data. It can be verified that this triplet ranking loss is convex, which can be easily integrated into the back propagation process of neural networks. 

\subsection{Horizontal pyramid}
This pyramid consists of three main building blocks: (1) the lateral hashing module to leverage pyramidal feature from different stage of CNN for hashing; (2) the consensus module to compose a consensus hash code from lateral hash feature. (3) the binary embedding module to map the consensus hash code to binary hash code. 

\subsubsection{\textbf{Lateral Hashing Module}}
In the proposed architecture for hashing, one of the key components is the lateral hashing module. The goal of this component is to transform the feature map into a specified dimension hash feature.  As shown in Figure \ref{Overview}, there are three connections between the two pyramids, and the lateral hashing module is based on the lateral connections. Therefore, we will present the details according to the lateral connection. First, for the middle lateral connections whose colour is black, as shown in the Figure \ref{Overview}, feature map $\mathbf{m}^{(s3)}$ is used as input to the lateral hash module. We first apply an average pooling layer $(Avgpool)$ to zoom out the feature map (i.e., a $2\times2$ or $4\times4$) before adding a fully connected layer on account of the fact that using a fully connected layer on too many elements may bring serious computational complexity. At the same time, zooming out the feature map is not directly discarding the spatial information in feature map on the same channel, for the fact that there are 4 and 16 spatial partitions on the feature maps of size $2\times2$ and $4\times4$, respectively. Moreover, reducing the size of the feature map does not affect the channel semantics information we use at different stages. Getting a relatively small feature map, we expand it and input it into the fully connected layer $(FC)$, which can be formulated as:
\begin{equation} \label{fc2}
  \begin{split}
  &\mathbf{a}^{(s3)} = Avgpool(\mathbf{m}^{(s3)}) ,    \\
  &\mathbf{f}^{(s3)} = FC(\mathbf{a}^{(s3)}) ,
  \end{split}
\end{equation} where $\mathbf{m}^{(s3)}\in{\mathbb{R}^{256\times{14}\times{14}}}$, $\mathbf{a}^{(s3)}\in{\mathbb{R}^{256\times{2}\times{2}}}$, and $\mathbf{f}^{(s2)}$ is $q\times2$-dimensional hash feature.

For the lower position lateral connections, we imitate the pipeline of the higher one:
\begin{equation} \label{fc3}
  \begin{split}
  &\mathbf{a}^{(s2)} = Avgpool(\mathbf{m}^{(s2)}) ,   \\
  &\mathbf{f}^{(s2)} = FC(\mathbf{a}^{(s2)}) ,
  \end{split}
\end{equation} where $\mathbf{m}^{(s2)}\in{\mathbb{R}^{128\times{28}\times{28}}}$, $\mathbf{a}^{(s2)}\in{\mathbb{R}^{128\times{4}\times{4}}}$, and $\mathbf{f}^{(s2)}$ is $q\times4$-dimensional hash feature.

In order to reduce the repeated calculation, for the hash feature of stage 4, we directly share the hash feature ${\mathbf{f}^{(s4)}}$ of the hashing module in the vertical pyramid. Specially, the design of the lateral hashing module is based on the structure of the hashing module in the vertical pyramid. As the annotation in equation (\ref{fc1}), (\ref{fc2}) and (\ref{fc3}) shown, the feature's dimension in lateral module actually follows a specific diminution pattern and incremental pattern. The feature map ${\mathbf{a}}$ is incremented according to the aspect ratio of the original feature map $\mathbf{m}$ (i.e., ${7\times7\xrightarrow{}1\times1 }$ correspond to ${ 14\times14\xrightarrow{}2\times2}$). With regard to incremental of the feature $\mathbf{f}$, considering that the low-stage feature map has too many elements, passing it directly through the fully-connected layer to obtain a feature vector with too low dimensions may result in excessive semantic loss and easy over-fitting \cite{overfit}. Therefore, the strategy we use is that the lower the feature map of the stage, the more feature elements are obtained through the fully connected layer. 
\subsubsection{\textbf{Consensus Module}}
Equipped with the pyramidal hash feature, we use two "mediator" to compose a consensus hash code. We first reveal the implementation details of the mediator $M1$ on the left-hand side. More specifically in $M1$, we apply an average pooling layer on $\mathbf{f}^{(s2)}$ to compress its dimensions from $q\times4$ to $q\times2$. And then we add it to the $\mathbf{f}^{(s3)}$ to get a new feature vector $\mathbf{f}^{(M1)}$ as the output of the mediator, which can be formulated as:
\begin{equation} 
  \begin{split}
  &\mathbf{f}^{(M1)} = Avgpool(\mathbf{f}^{(s2)}) + \mathbf{f}^{(s3)} ,
  \end{split}
\end{equation} where $\mathbf{f}^{(M1)}$ is $q\times2$-dimensional hash feature. 

Considering that the hash code we used for retrieval is $q$ dimension, we need average pooling layer to compress the dimension of the low-stage hash feature. The same strategy applies to the design of the mediator $M2$ on the right-hand side, and it can also be formulated as:
\begin{equation}
  \begin{split}
  &\mathbf{f}^{(M2)} = Avgpool(\mathbf{f}^{(M1)}) + \mathbf{f}^{(s4)} ,
  \end{split}
\end{equation} where $\mathbf{f}^{(M2)}$ is $q$-dimensional consensus hash feature. 

Considering the rarity of hash code for retrieval, we do not directly fix some bits with the feature of certain stage but adopt a fusion method to generate a consensus of several stages as a hash code on each bit of hash vector. For the two mediators, we combine the features of different stages, so that each bit of the final output $\mathbf{f}^{(M2)}$ is not determined by a certain layer alone, but determined by several stages' "comments" to produce a consensus on the final hash feature.

With the $q$-dimensional consensus hash feature, we employ a sigmoid activation layer $(Sigmoid)$ to restrict each element  of the hash feature to the range [0, 1]. We denote the output vector of $(Sigmoid)$ as $\mathbf{v}^{c}$:
\begin{equation}
  \begin{split}
  &\mathbf{v}^{c} = Sigmoid(\mathbf{f}^{(M2)}) ,
  \end{split}
\end{equation} where $\mathbf{v}^{c}$ is $q$-dimensional consensus hash code. 

In the end, we still use the triplet ranking loss to preserve the semantic similarity between different images with the consensus hash code. As with the triplet ranking loss detailed above, for a triplet of images $(I_{i},I_{j},I_{k})$ that $I_{i}$ is more similar to $I_{j}$ than to $I_{k}$, we denote the the real-valued consensus hash code for $I_{i}$, $I_{j}$ and $I_{k}$ as ${\mathbf{v_i}^{c}}$, ${\mathbf{v_j}^{c}}$ and ${\mathbf{v_k}^{c}}$, respectively. And the loss function in the consensus module can be defined as:
\begin{equation}\label{loss}
\begin{split}
&\ell_{tri}({\mathbf{v_i}^{c}},\mathbf{v_j}^{c},\mathbf{v_k}^{c})\\
=&\max(0,m_n+||{\mathbf{v_i}^{c}}-\mathbf{v_j}^{c}||_2^2-||{\mathbf{v_i}^{c}}-\mathbf{v_k}^{c}||_2^2)\\
&s.t.\ {\mathbf{v_i}^{c}},\ \mathbf{v_j}^{c},\ \mathbf{v_k}^{c}
\in [0,1]^q,\\
\end{split}
\end{equation}where $m_n$ is the margin parameter depending on the hash code length $q$, $||.||_2$ is the $\ell_2$ norm.
\\
\textbf{Combination of Loss Functions}
During the training phase, we use a combination of the above loss functions with stochastic gradient descent defined by:
\begin{equation} \label{multi-task loss}
\begin{split}
\ell_{comb} = &\frac{1}{M} \sum_{i=1}^{M}( \ell_{tri}(\mathbf{v_i}, \mathbf{v_j}, \mathbf{v_k})+\ell_{tri}(\mathbf{v_i}^{c}, \mathbf{v_j}^{c}, \mathbf{v_k}^{c})),\\
\end{split}
\end{equation} where $(\mathbf{v_i}, \mathbf{v_j}, \mathbf{v_k})$ and $(\mathbf{v_i}^{c}, \mathbf{v_j}^{c}, \mathbf{v_k}^{c})$ are the hash code and consensus hash code corresponding to the triplet images $(I_i, I_j, I_k)$, M is the number of triplets.

\subsubsection{\textbf{Binary Embedding Module}}
This module mainly works in the test phase. Specifically, for an input image $I$ and its consensus hash code $\mathbf{v}^{c}$, the $q$-bit binary code b can be obtained by:
\begin{equation}
\label{threshold}
b_{i}=\left\{
\begin{aligned}
  1, \quad & v_{i}^{c} \geq 0.5 \\
  0, \quad & {v_i}^{c}< 0.5,
\end{aligned}
\right.
\end{equation}where $b_i$/${v_i}^{c}$ is the $i$-th element in $\mathbf{b}$/$\mathbf{v}^{c}$, respectively.


\begin{table*}[htbp]
\caption{MAP of Hamming ranking w.r.t different number of bits on two fine-grained datasets.}
\label{MAP}
\begin{tabular}{|c|
p{1.2cm}<{\centering}p{1.2cm}<{\centering}p{1.2cm}<{\centering}p{1.2cm}<{\centering}|
p{1.2cm}<{\centering}p{1.2cm}<{\centering}p{1.2cm}<{\centering}p{1.2cm}<{\centering}|
}
\hline
\multirow{2}{*}{{Methods}}& \multicolumn{4}
{c|}{CUB-200-2011}&\multicolumn{4}
{c|}{Stanford Dogs}\\
& 16bits & 32bits & 48bits & 64bits 
& 16bits & 32bits & 48bits & 64bits 
\\
\hline
{\bf Ours}&  {\bf 0.5169}&  {\bf 0.5832}&  {\bf 0.6124}& {\bf 0.6233}& {\bf 0.6340}&  {\bf 0.6909}&  {\bf 0.7060}& {\bf 0.7130}\\ \hline

DTH \cite{lai2015simultaneous}&  0.4641&  0.5454&  0.5771&  0.5881&    0.5435&  0.6258&  0.6362&  0.6573\\  \hline

DSH \cite{liu2016deep}&  0.3156&  0.4930&  0.5408&  0.5967&    0.4728&  0.5587&  0.6128&  0.6319\\  \hline 
HashNet \cite{cao2017hashnet}&  0.3791&  0.4628&  0.4853&  0.5123&    0.4745&  0.5521&  0.5575&  0.5934\\  \hline

DPSH \cite{Li2016Feature}&  0.3497&  0.4301&  0.4908&  0.5225&    0.4270&  0.5528&  0.6080&  0.6231\\  \hline \hline

CCA-ITQ&      0.1142&  0.1580&  0.1813&  0.1986&    0.2632&  0.3681&  0.4175&  0.4402\\  \hline

MLH&      0.0915&  0.1289&  0.1281&  0.1983&    0.2735&  0.3531&  0.3831&  0.4084\\  \hline

ITQ&      0.0637&  0.0907&  0.1048&  0.1129&    0.2023&  0.2838&  0.3123&  0.3248\\  \hline
SH&      0.0453&  0.0595&  0.0643&  0.0686&    0.1362&  0.1628&  0.1859&  0.1832\\  \hline
LSH&      0.0162&  0.0234&  0.0302&  0.0340&    0.0297&  0.0517&  0.0640&  0.0850\\  \hline

\end{tabular}
\end{table*}

\begin{figure*}[ht!]
\centering
\subfigure[Precision within Hamming radius 3]
{
\label{topp:subfig:a}
\includegraphics[width=2.15in]{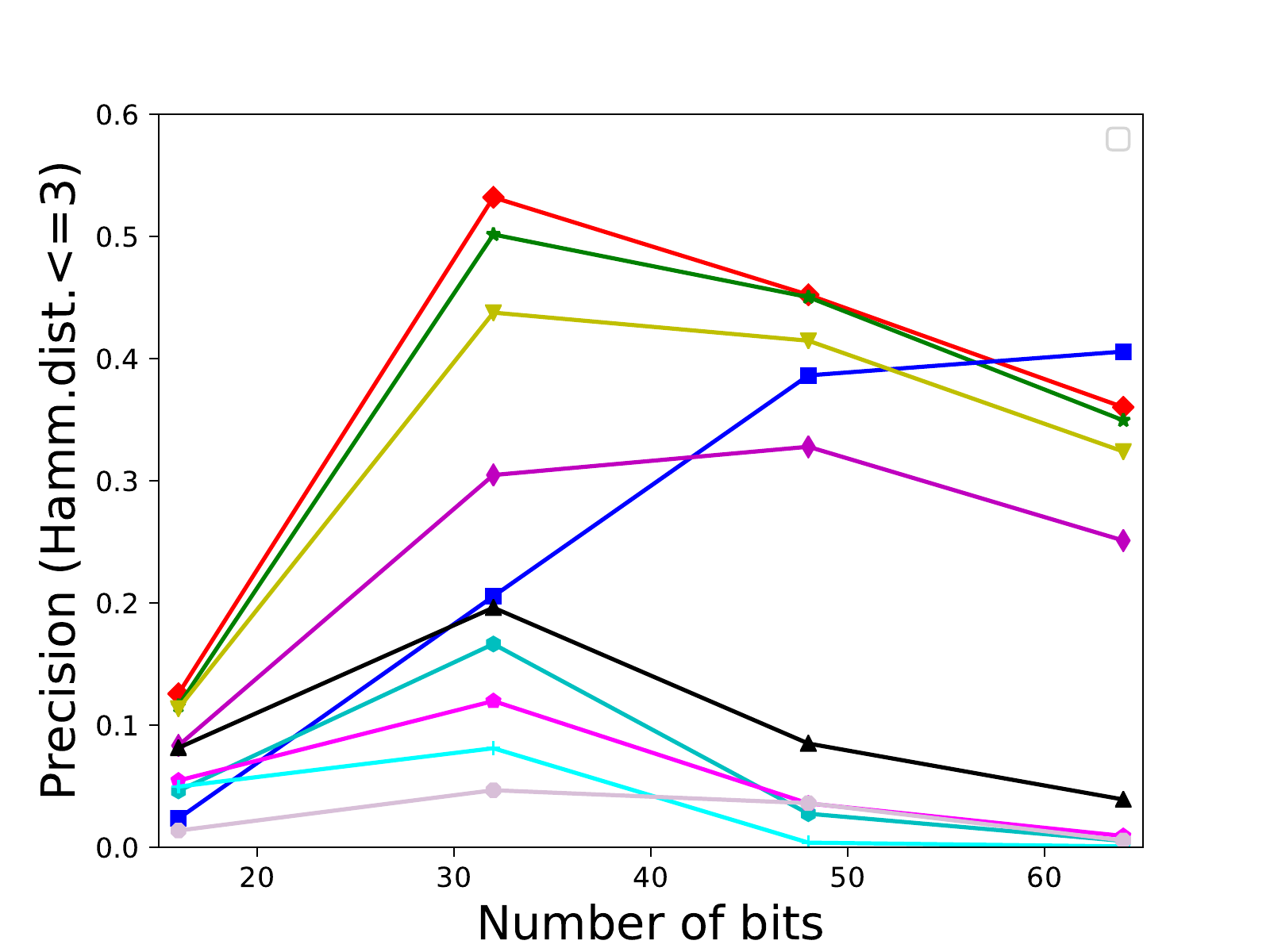}
}
\hspace{0.0in}
\subfigure[Precision-recall curve @ 16 bits]
{
\label{topp:subfig:b}
\includegraphics[width=2.15in]{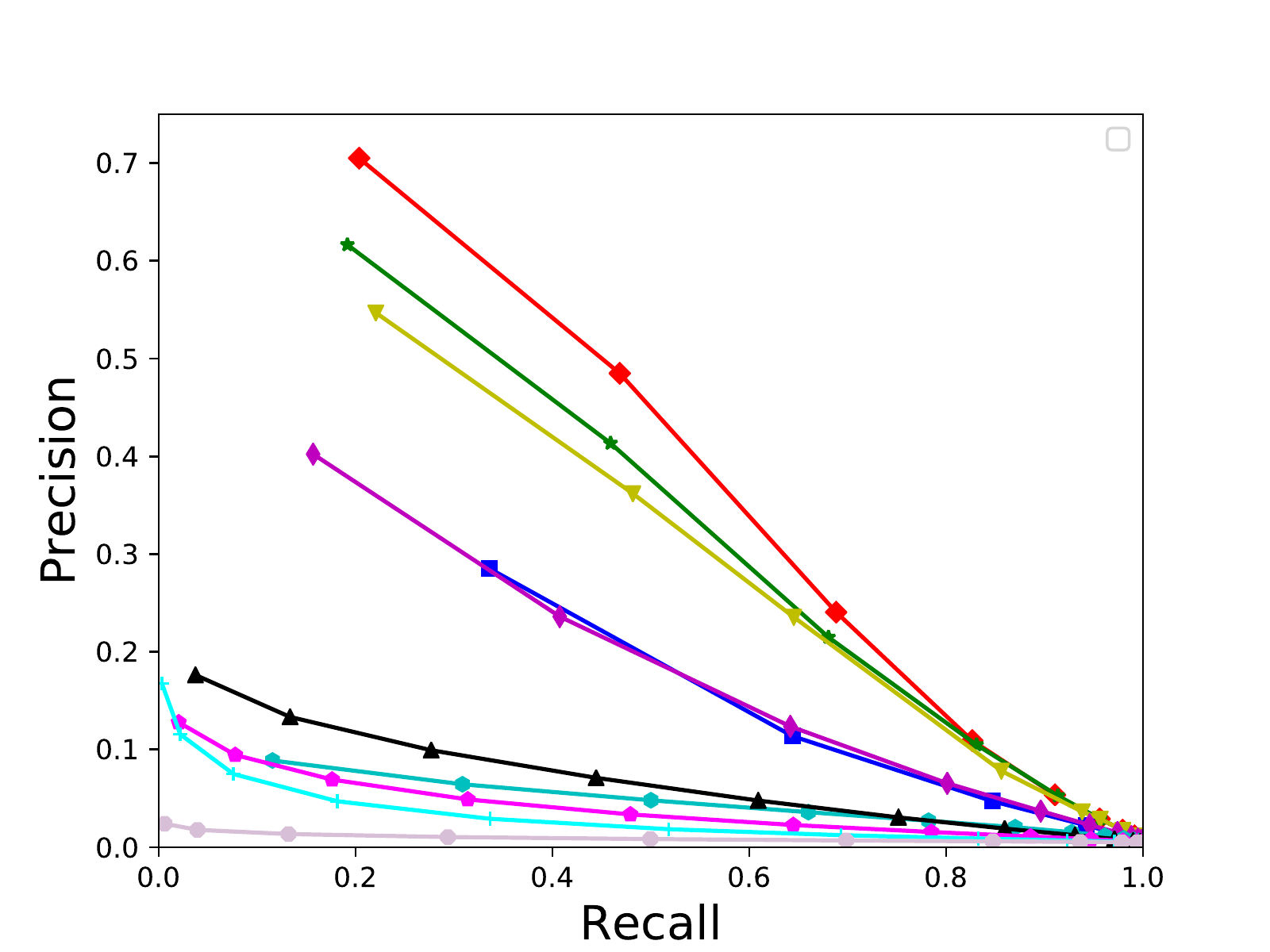}
}
\hspace{0.0in}
\subfigure[Precision curve w.r.t. top-N @ 16 bits]
{
\label{topp:subfig:b}
\includegraphics[width=2.15in]{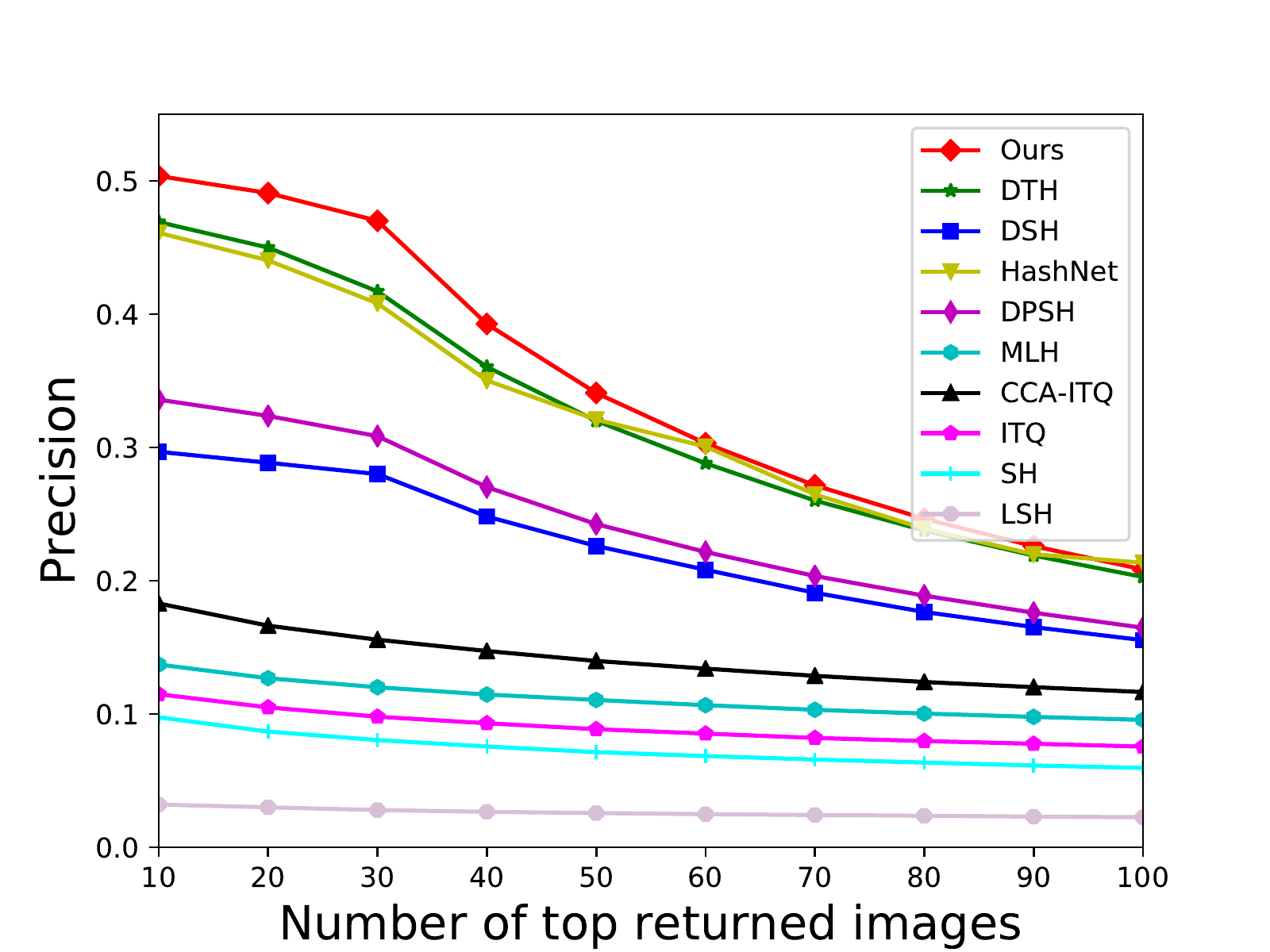}
}
\caption{The experimental results on the CUB-200-2011 dataset under three evaluation metrics.}
\label{topp1}
\end{figure*}

\begin{figure*}[ht!]
\centering
\subfigure[Precision within Hamming radius 3]
{
\label{topp:subfig:a}
\includegraphics[width=2.15in]{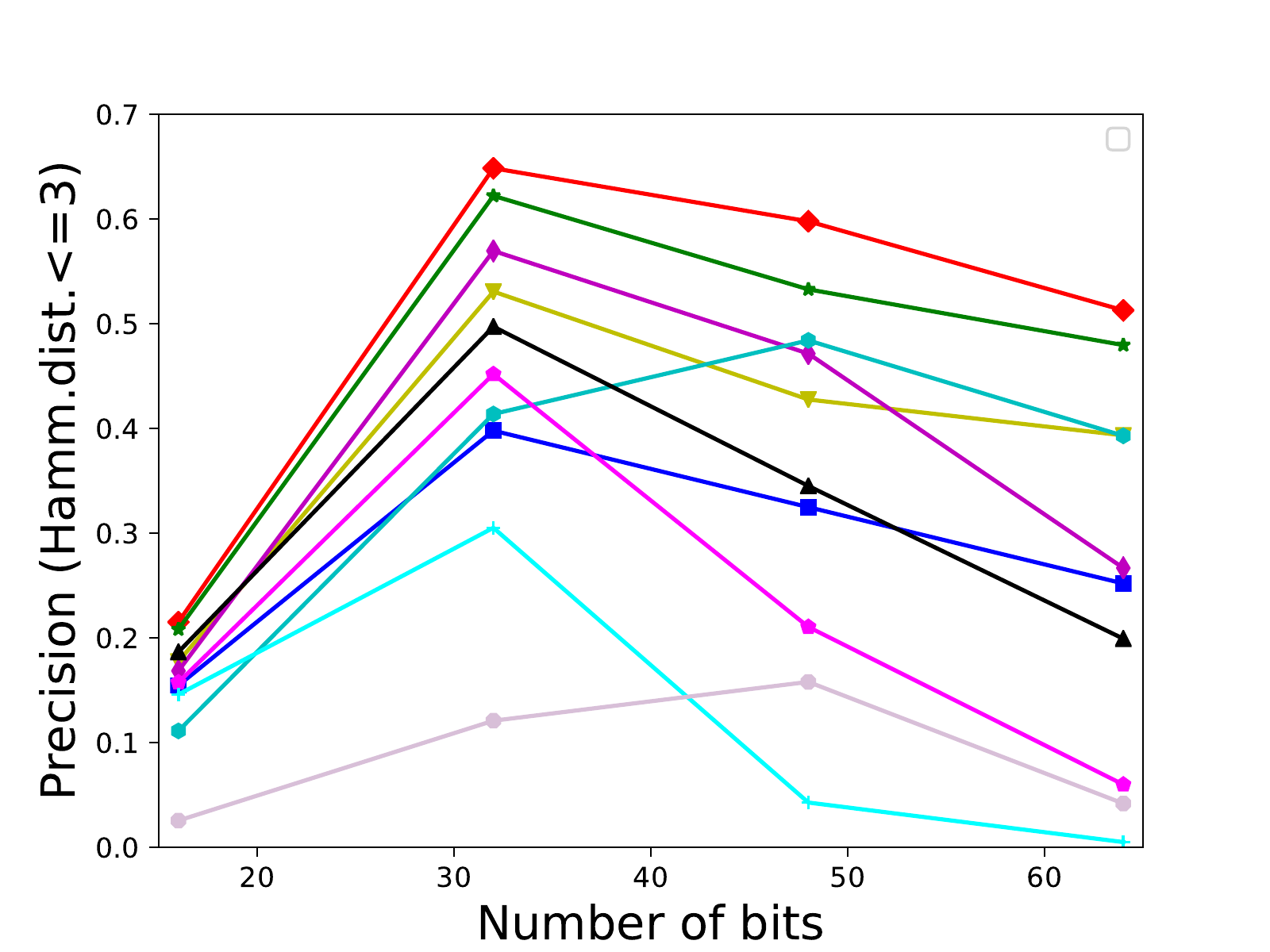}
}
\hspace{0.0in}
\subfigure[Precision-recall curve @ 16 bits]
{
\label{topp:subfig:b}
\includegraphics[width=2.15in]{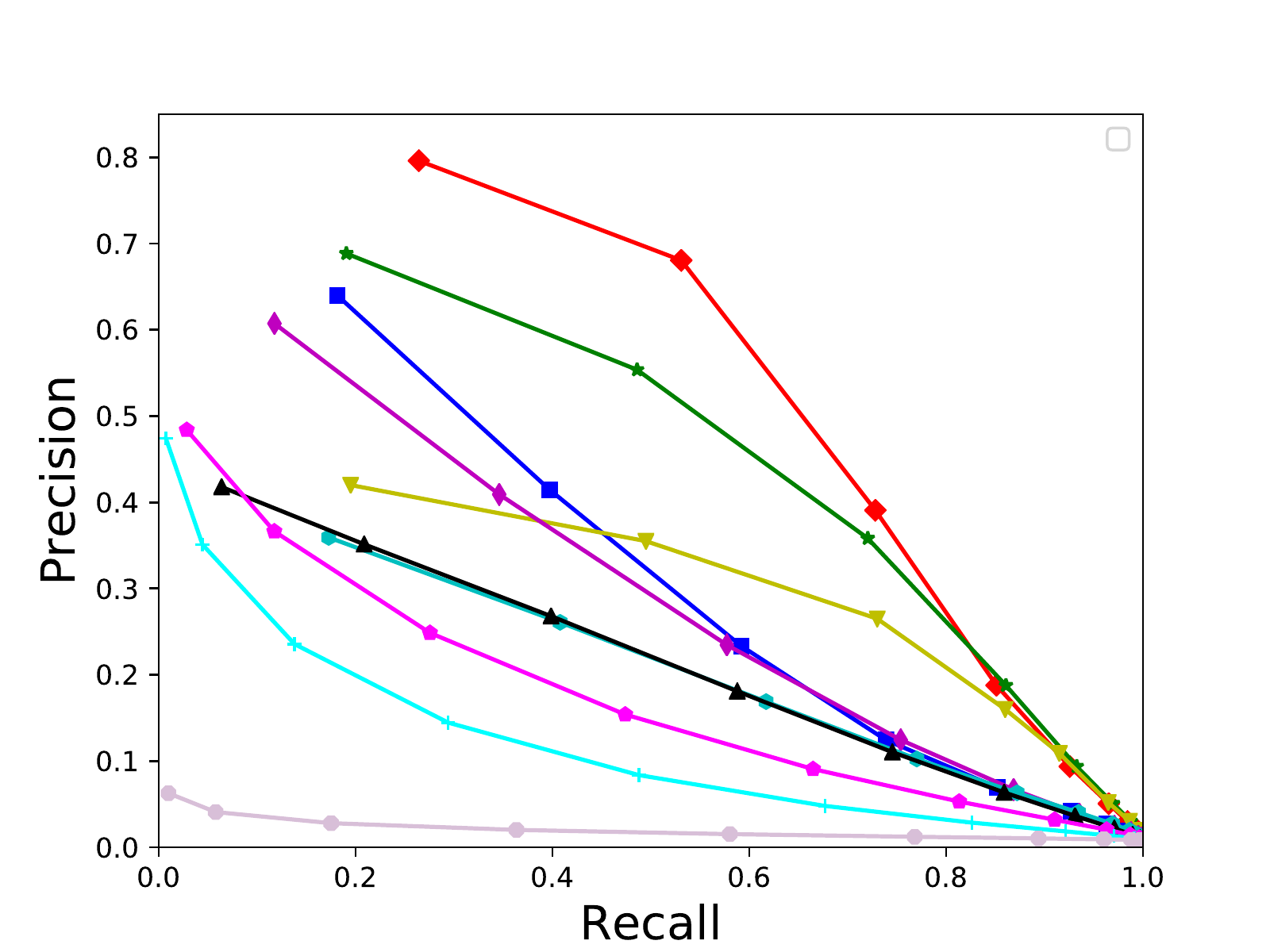}
}
\hspace{0.0in}
\subfigure[Precision curve w.r.t. top-N @ 16 bits]
{
\label{topp:subfig:b}
\includegraphics[width=2.15in]{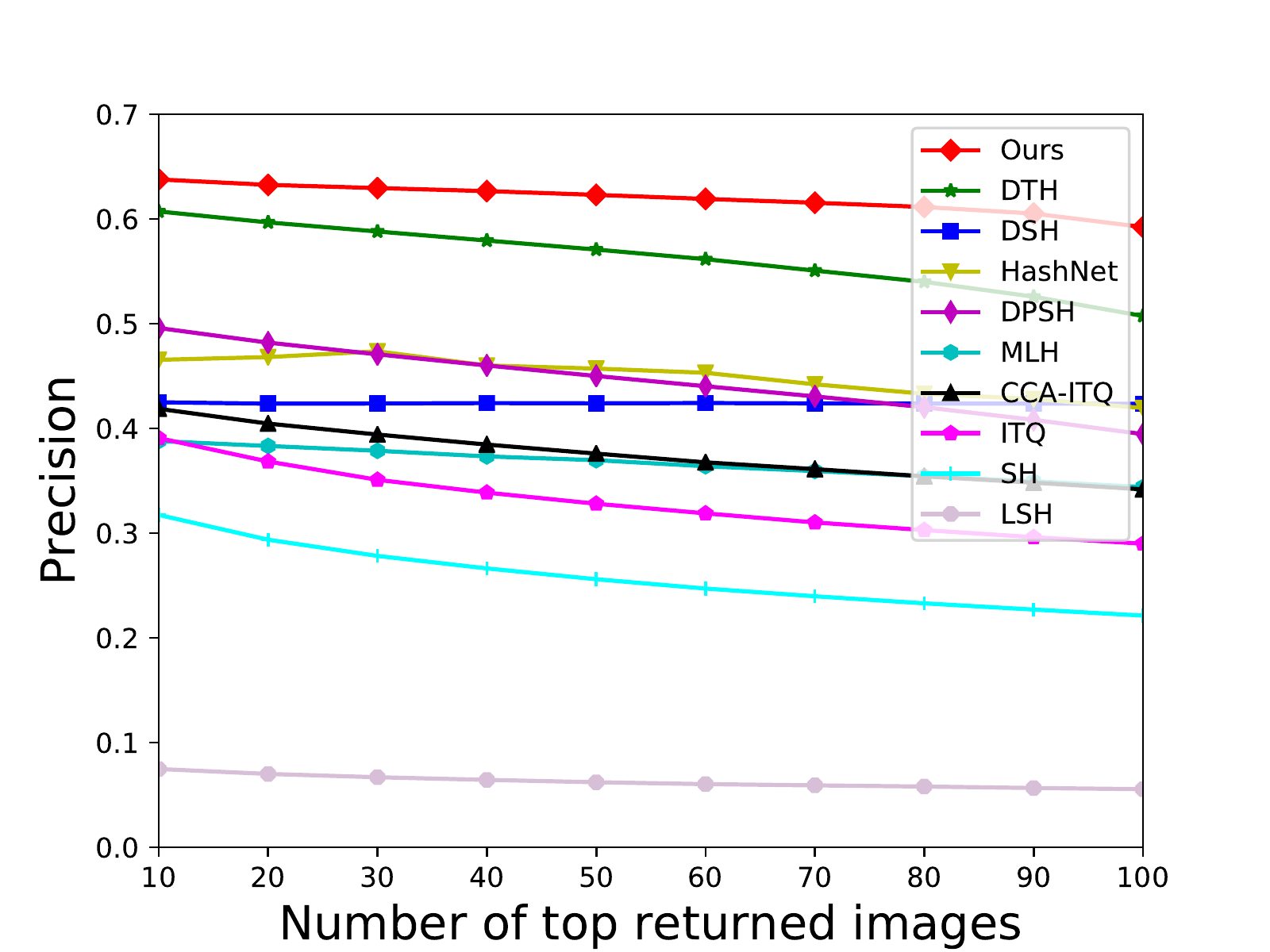}
}
\caption{The experimental results on the Standford Dogs dataset under three evaluation metrics.}
\label{topp2}
\end{figure*}

\begin{table*}[htbp]
\caption{Comparison with DaSH\cite{jin2018deep} of MAP on two fine-grained datasets.}
\label{MAP_Dash}
\begin{tabular}{|c|
p{1.2cm}<{\centering}p{1.2cm}<{\centering}p{1.2cm}<{\centering}p{1.2cm}<{\centering}|
p{1.2cm}<{\centering}p{1.2cm}<{\centering}p{1.2cm}<{\centering}p{1.2cm}<{\centering}|
}
\hline
\multirow{2}{*}{{Methods}}& \multicolumn{4}
{c|}{Oxford Flower-17}&\multicolumn{4}
{c|}{Stanford Dogs}\\
& 16bits & 32bits & 48bits & 64bits 
& 16bits & 32bits & 48bits & 64bits 
\\
\hline
{\bf Ours}&  {\bf 0.9542}&  {\bf 0.9653}&  {\bf 0.9691}& {\bf 0.9783}& {\bf 0.6224}&  {\bf 0.6688}&  {\bf 0.6924}& {\bf 0.6974}\\ \hline

DaSH \cite{jin2018deep}&  0.9225&  0.9267&  0.9692&  0.9756&    0.3976&  0.5283&  0.5950&  0.6452\\  \hline

\end{tabular}
\end{table*}

\section{Experiments}
\subsection{Datasets}
We conduct extensive evaluations of the proposed method and compare with state-of-the-art baselines on two fine-grained datasets: 
\begin{itemize}
\item \textbf{CUB-200-2011}\footnote{http://www.vision.caltech.edu/visipedia/CUB-200-2011.html}: It is an extended version of CUB-200 \cite{welinder2010caltech}, a challenging dataset that pushes the limits of visual abilities for both humans and computer consists of 11788 images of birds in 200 classes.  
\item \textbf{Standford Dogs}\footnote{http://vision.stanford.edu/aditya86/ImageNetDogs/}: It is a challenging and large-scale dataset which includes over 22,580 annotated images of dogs belonging to 120 species aimed at fine-grained image tasks. This dataset is extremely challenging due to two reason: first, there is little inter-class variation; second, there is very large intra-class variation.
\end{itemize}

In CUB-200-2011, we use the official split, where 5794 test images as the test query set, 5994 training images as the training set to train the hash models and also used as the retrieval database. In Standford Dogs, the official split will also be applied, 22,580 annotated images will be split into two parts, 12000 training samples and 8580 testing samples. All the training samples also serve as a retrieval database besides training the network. For a fair comparison, all of the methods for comparison use identical training/test sets and retrieval database. 

\subsection{Evaluation Metrics}
To measure the performance of hashing, we use four evaluation metrics: mean average precision(MAP), precision-recall curves, precision curve within hamming radius 3 and precision curves w.r.t. different numbers of top returned samples. MAP is a widely used evaluation measure for ranking, The average precision of image $x_i$ can be defined as:
\begin{equation} \label{AP}
\begin{split}
AP_{i} = &\frac{1}{N_+} \sum_{k=1}^{n}\frac{N_+^k}{k}\times pos(k)\\
\end{split}
\end{equation} where $n$ is the number of images in the retrieval database. The $pos(k)$ is an indicator function, in which if the image at position $k$ is positive, then $pos(k) = 1$, otherwise $pos(k) = 0$. The $N^k_+$ is the number of relevant images within the top $k$ images and $N_+$ represents the total number of relevant images w.r.t the $i$-th query image. For all query images, the MAP is defined as:
\begin{equation} 
\begin{split}
MAP = &\frac{1}{n_q} \sum_{i=1}^{n_q}AP_i\\
\end{split}
\end{equation}

\subsection{Settings and Implementation Details}
The experiments for our proposed method are completed with the open source PyTorch \cite{paszke2017pytorch} framework on a GeForce GTX TITAN X server.

For fair comparison, all deep CNN-based methods, including ours and previous baselines, are based on the same CNN architecture, i.e., ResNet \cite{resnet}. Specially, we remove the last fully-connected layer since it is for 1,000 classifications to make the rest of the ResNet act as the convolutional sub-network in our architecture. The weights of the convolutional sub-network are initialized with the pre-trained ResNet model \footnote{https://download.pytorch.org/models/resnet18-5c106cde.pth} that learns from the ImageNet dataset. 
For other non-deep-network-based methods, we use the pre-trained ResNet model to extract features from raw images and then use these features as input, i.e., the last layer of ResNet output 512-dimensional vector after removing the fully-connected layer from the pre-trained model.

For all the method, we resize all of the images to the size $224\times224$ and use the raw image pixels as input. During the training phase, all the training samples will divide into mini batches before inputting to the network and the batch size is 100. The proposed architecture in this paper is trained by the stochastic gradient descent with 0.9 momentum and 0.0005 weight decay for the triplet ranking loss function is a strictly convex function. The base learning rate is 0.001 and the step size is 1800 which means the learning rate will be 10 times smaller every 1800 epoch where the total epoch is set to be 4000. 

We compare the proposed method with several state-of-the-art learning-based hashing methods, which can be roughly divided into two categorized:
\begin{itemize}
\item Conventional hashing methods: ITQ\cite{gong2013iterative} reduced the quantization errors by learning an orthogonal rotation matrix; CCA-ITQ \cite{yunchao2013iterative}, an extension of ITQ, uses label information to find better projections for the image descriptors; LSH \cite{gionis1999similarity} uses random projections to produce hashing bits. Spectral Hashing (SH) \cite{weiss2009spectral} tries to minimizes the weighted Hamming distance of image pairs, where the weights are defined to be the similarity metrics of image pairs. MLH \cite{norouzi2011minimal} uses structural SVMs with latent variables to encodes images. 
\item Deep-network-based hashing methods including: DTH \cite{lai2015simultaneous} proposed a deep triplet-based loss function for supervised hashing method. DPSH \cite{li2015feature} proposed a deep hashing method to perform simultaneous feature learning and hash code learning for applications with pairwise labels. DSH \cite{liu2016deep} speeds up the training of the network by adding a regular term to loss function instead of using activation function and employs the pairwise loss function with margin to preserve the similarity of images. HASHNET \cite{cao2017hashnet} utilizes the weighted pairwise loss to maximize the WML likelihood function, and takes a weighted attenuation factor on the activation function, thereby reducing the semantic loss caused by feature-to-hash code mapping. Additionally, for hashing method designed specifically for fine-grained data, deep saliency hashing(DSaH) \cite{jin2018deep} uses an attention mechanism to learn the hashing codes. 
\end{itemize}

Specially, the implementation of DTH in this paper is a variant of \cite{lai2015simultaneous}, in which we replace the divide-and-encode module by a fully connected layer with sigmoid activation. Note that the architecture of DTH is just the same as that of the vertical pyramid without side output, which makes it convenient for us to observe the improvement made by the horizontal pyramid.

\subsection{Experimental Results}
\subsubsection{Comparison with State-of-the-art Methods}
To illustrate the accuracy of the proposed method, we evaluate and compare our method with several state-of-the-art baselines.

Compared to other hashing baselines, the proposed method shows substantially better performance gains. Take our main evaluation metrics MAP as an example, as shown in Table \ref{MAP}, the proposed method shows a relative improvement of a relative improvement of \textbf{5.9$\%\sim$11.3$\%$}/\textbf{8.4$\%\sim$16.6$\%$} against the second best baseline on  CUB-200-2011/Stanford  Dogs, respectively. In addition, in Figure \ref{topp1} and Figure \ref{topp2}, it can be observed that the proposed method performs better than all previous methods in precision with hamming radius 3 curves, precision-recall and precision on 16bits for most levels. 

In particular, the proposed method consistently outperforms DTH. Since the implementation of the DTH is exactly the same as that of the vertical pyramid in Figure \ref{Overview} (a), DTH is equivalent to the proposed method without the pyramid architecture using the pyramidal feature to compose the hash code. The predominant performance of the proposed method against DTH verifies that using the pyramidal feature consensus to compose the hash code can improve the performance of deep hashing. 

\subsubsection{Comparison with Hashing Method Specifically Designed for Fine-grained Data}
As our approach works primarily for fine-grained data, we compare it to other approaches specifically designed for fine-grained data.

Since the code of DaSH\cite{jin2018deep} is not publicly available, and it is hard to re-implement the complex method, we utilize the same experimental settings used in DaSH for our method. The results of DaSH are directly cited from \cite{jin2018deep} for a fair comparison. Following the DaSH setting, we also use the VGG \cite{simonyan2014very} as the basic architecture, which has the same expression ability as ResNet.

In DaSH\cite{jin2018deep}, two datasets are mainly used: the first is the Stanford Dogs which is described in detail above, and the second is the Oxford Flower-17\footnote{http://www.robots.ox.ac.uk/~vgg/data/flowers/17/}, which consists of dataset consists of 1360 images of flowers belonging to 17 mutually classes. The training set and the test set partition also follow the setting in DaSH.

The MAP results on Oxford Flower-17 and Stanford Dogs are shown in Table \ref{MAP_Dash}, which show the superior performance gain of the proposed method over the other approach specifically designed for fine-grained data. 






\section{Conclusions}
In this paper, we developed a clean and simple two-pyramid architecture that learning both the semantic information and the subtle appearance details from fine-grained objects to improve the performance of deep hashing, in which the vertical pyramid capture the high-layer features and the horizontal pyramid combines multiple low-layer features with more structural information to capture the subtle differences. Empirical evaluations on two representative fine-grained images datasets show that the proposed method achieves better performance of deep hashing.

\bibliographystyle{ACM-Reference-Format}
\bibliography{acmart}


\begin{thebibliography}{40}


\ifx \showCODEN    \undefined \def \showCODEN     #1{\unskip}     \fi
\ifx \showDOI      \undefined \def \showDOI       #1{#1}\fi
\ifx \showISBNx    \undefined \def \showISBNx     #1{\unskip}     \fi
\ifx \showISBNxiii \undefined \def \showISBNxiii  #1{\unskip}     \fi
\ifx \showISSN     \undefined \def \showISSN      #1{\unskip}     \fi
\ifx \showLCCN     \undefined \def \showLCCN      #1{\unskip}     \fi
\ifx \shownote     \undefined \def \shownote      #1{#1}          \fi
\ifx \showarticletitle \undefined \def \showarticletitle #1{#1}   \fi
\ifx \showURL      \undefined \def \showURL       {\relax}        \fi
\providecommand\bibfield[2]{#2}
\providecommand\bibinfo[2]{#2}
\providecommand\natexlab[1]{#1}
\providecommand\showeprint[2][]{arXiv:#2}

\bibitem[\protect\citeauthoryear{Cao, Long, Liu, Wang, and KLiss}{Cao
  et~al\mbox{.}}{2018}]%
        {cao2018deep}
\bibfield{author}{\bibinfo{person}{Yue Cao}, \bibinfo{person}{Mingsheng Long},
  \bibinfo{person}{Bin Liu}, \bibinfo{person}{Jianmin Wang}, {and}
  \bibinfo{person}{MOE KLiss}.} \bibinfo{year}{2018}\natexlab{}.
\newblock \showarticletitle{Deep Cauchy Hashing for Hamming Space Retrieval}.
  In \bibinfo{booktitle}{\emph{Proceedings of the IEEE Conference on Computer
  Vision and Pattern Recognition}}. \bibinfo{pages}{1229--1237}.
\newblock


\bibitem[\protect\citeauthoryear{Cao, Long, Wang, and Philip}{Cao
  et~al\mbox{.}}{2017}]%
        {cao2017hashnet}
\bibfield{author}{\bibinfo{person}{Zhangjie Cao}, \bibinfo{person}{Mingsheng
  Long}, \bibinfo{person}{Jianmin Wang}, {and} \bibinfo{person}{S~Yu Philip}.}
  \bibinfo{year}{2017}\natexlab{}.
\newblock \showarticletitle{HashNet: Deep Learning to Hash by Continuation.}.
  In \bibinfo{booktitle}{\emph{Proceedings of the IEEE international conference
  on computer vision}}. \bibinfo{pages}{5609--5618}.
\newblock


\bibitem[\protect\citeauthoryear{Chen, Wang, Peng, Zhang, Yu, and Sun}{Chen
  et~al\mbox{.}}{2018}]%
        {chen2018cascaded}
\bibfield{author}{\bibinfo{person}{Yilun Chen}, \bibinfo{person}{Zhicheng
  Wang}, \bibinfo{person}{Yuxiang Peng}, \bibinfo{person}{Zhiqiang Zhang},
  \bibinfo{person}{Gang Yu}, {and} \bibinfo{person}{Jian Sun}.}
  \bibinfo{year}{2018}\natexlab{}.
\newblock \showarticletitle{Cascaded pyramid network for multi-person pose
  estimation}. In \bibinfo{booktitle}{\emph{Proceedings of the IEEE Conference
  on Computer Vision and Pattern Recognition}}. \bibinfo{pages}{7103--7112}.
\newblock


\bibitem[\protect\citeauthoryear{Gionis, Indyk, Motwani, et~al\mbox{.}}{Gionis
  et~al\mbox{.}}{1999}]%
        {gionis1999similarity}
\bibfield{author}{\bibinfo{person}{Aristides Gionis}, \bibinfo{person}{Piotr
  Indyk}, \bibinfo{person}{Rajeev Motwani}, {et~al\mbox{.}}}
  \bibinfo{year}{1999}\natexlab{}.
\newblock \showarticletitle{Similarity search in high dimensions via hashing}.
  In \bibinfo{booktitle}{\emph{Proceedings of the International Conference on
  Very Large Data Bases}}, Vol.~\bibinfo{volume}{99}.
  \bibinfo{pages}{518--529}.
\newblock


\bibitem[\protect\citeauthoryear{Gong, Lazebnik, Gordo, and Perronnin}{Gong
  et~al\mbox{.}}{2013}]%
        {gong2013iterative}
\bibfield{author}{\bibinfo{person}{Yunchao Gong}, \bibinfo{person}{Svetlana
  Lazebnik}, \bibinfo{person}{Albert Gordo}, {and} \bibinfo{person}{Florent
  Perronnin}.} \bibinfo{year}{2013}\natexlab{}.
\newblock \showarticletitle{Iterative quantization: A procrustean approach to
  learning binary codes for large-scale image retrieval}.
\newblock \bibinfo{journal}{\emph{IEEE Transactions on Pattern Analysis and
  Machine Intelligence}} \bibinfo{volume}{35}, \bibinfo{number}{12}
  (\bibinfo{year}{2013}), \bibinfo{pages}{2916--2929}.
\newblock


\bibitem[\protect\citeauthoryear{He, Zhang, Ren, and Sun}{He
  et~al\mbox{.}}{2016a}]%
        {resnet}
\bibfield{author}{\bibinfo{person}{Kaiming He}, \bibinfo{person}{Xiangyu
  Zhang}, \bibinfo{person}{Shaoqing Ren}, {and} \bibinfo{person}{Jian Sun}.}
  \bibinfo{year}{2016}\natexlab{a}.
\newblock \showarticletitle{Deep residual learning for image recognition}. In
  \bibinfo{booktitle}{\emph{Proceedings of the IEEE conference on computer
  vision and pattern recognition}}. \bibinfo{pages}{770--778}.
\newblock


\bibitem[\protect\citeauthoryear{He, Zhang, Ren, and Sun}{He
  et~al\mbox{.}}{2016b}]%
        {overfit}
\bibfield{author}{\bibinfo{person}{Kaiming He}, \bibinfo{person}{Xiangyu
  Zhang}, \bibinfo{person}{Shaoqing Ren}, {and} \bibinfo{person}{Jian Sun}.}
  \bibinfo{year}{2016}\natexlab{b}.
\newblock \showarticletitle{Deep residual learning for image recognition}. In
  \bibinfo{booktitle}{\emph{Proceedings of the IEEE conference on computer
  vision and pattern recognition}}. \bibinfo{pages}{770--778}.
\newblock


\bibitem[\protect\citeauthoryear{Jin}{Jin}{2018}]%
        {jin2018deep}
\bibfield{author}{\bibinfo{person}{Sheng Jin}.}
  \bibinfo{year}{2018}\natexlab{}.
\newblock \showarticletitle{Deep Saliency Hashing}.
\newblock \bibinfo{journal}{\emph{arXiv preprint arXiv:1807.01459}}
  (\bibinfo{year}{2018}).
\newblock


\bibitem[\protect\citeauthoryear{Kong, Sun, Tan, Liu, and Huang}{Kong
  et~al\mbox{.}}{2018}]%
        {kong2018deep}
\bibfield{author}{\bibinfo{person}{Tao Kong}, \bibinfo{person}{Fuchun Sun},
  \bibinfo{person}{Chuanqi Tan}, \bibinfo{person}{Huaping Liu}, {and}
  \bibinfo{person}{Wenbing Huang}.} \bibinfo{year}{2018}\natexlab{}.
\newblock \showarticletitle{Deep feature pyramid reconfiguration for object
  detection}. In \bibinfo{booktitle}{\emph{Proceedings of the European
  Conference on Computer Vision}}. \bibinfo{pages}{169--185}.
\newblock


\bibitem[\protect\citeauthoryear{Kulis and Darrell}{Kulis and Darrell}{2009}]%
        {kulis2009learning}
\bibfield{author}{\bibinfo{person}{Brian Kulis} {and} \bibinfo{person}{Trevor
  Darrell}.} \bibinfo{year}{2009}\natexlab{}.
\newblock \showarticletitle{Learning to hash with binary reconstructive
  embeddings}. In \bibinfo{booktitle}{\emph{Advances in neural information
  processing systems}}. \bibinfo{pages}{1042--1050}.
\newblock


\bibitem[\protect\citeauthoryear{Lai, Pan, Liu, and Yan}{Lai
  et~al\mbox{.}}{2015}]%
        {lai2015simultaneous}
\bibfield{author}{\bibinfo{person}{Hanjiang Lai}, \bibinfo{person}{Yan Pan},
  \bibinfo{person}{Ye Liu}, {and} \bibinfo{person}{Shuicheng Yan}.}
  \bibinfo{year}{2015}\natexlab{}.
\newblock \showarticletitle{Simultaneous feature learning and hash coding with
  deep neural networks}. In \bibinfo{booktitle}{\emph{Proceedings of the IEEE
  conference on computer vision and pattern recognition}}.
  \bibinfo{pages}{3270--3278}.
\newblock


\bibitem[\protect\citeauthoryear{Li, Wang, and Kang}{Li et~al\mbox{.}}{2015}]%
        {li2015feature}
\bibfield{author}{\bibinfo{person}{Wu-Jun Li}, \bibinfo{person}{Sheng Wang},
  {and} \bibinfo{person}{Wang-Cheng Kang}.} \bibinfo{year}{2015}\natexlab{}.
\newblock \showarticletitle{Feature learning based deep supervised hashing with
  pairwise labels}.
\newblock \bibinfo{journal}{\emph{arXiv preprint arXiv:1511.03855}}
  (\bibinfo{year}{2015}).
\newblock


\bibitem[\protect\citeauthoryear{Li, Wang, and Kang}{Li et~al\mbox{.}}{2016}]%
        {Li2016Feature}
\bibfield{author}{\bibinfo{person}{Wu~Jun Li}, \bibinfo{person}{Sheng Wang},
  {and} \bibinfo{person}{Wang~Cheng Kang}.} \bibinfo{year}{2016}\natexlab{}.
\newblock \showarticletitle{Feature learning based deep supervised hashing with
  pairwise labels}. In \bibinfo{booktitle}{\emph{Proceedings of the
  International Joint Conference on Artificial Intelligence}}.
  \bibinfo{pages}{1711--1717}.
\newblock


\bibitem[\protect\citeauthoryear{Lin, Shen, Suter, and Van Den~Hengel}{Lin
  et~al\mbox{.}}{2013}]%
        {lin2013general}
\bibfield{author}{\bibinfo{person}{Guosheng Lin}, \bibinfo{person}{Chunhua
  Shen}, \bibinfo{person}{David Suter}, {and} \bibinfo{person}{Anton Van
  Den~Hengel}.} \bibinfo{year}{2013}\natexlab{}.
\newblock \showarticletitle{A general two-step approach to learning-based
  hashing}. In \bibinfo{booktitle}{\emph{Proceedings of the IEEE international
  conference on computer vision}}. \bibinfo{pages}{2552--2559}.
\newblock


\bibitem[\protect\citeauthoryear{Lin, Lu, Chen, and Zhou}{Lin
  et~al\mbox{.}}{2016}]%
        {lin2016learning}
\bibfield{author}{\bibinfo{person}{Kevin Lin}, \bibinfo{person}{Jiwen Lu},
  \bibinfo{person}{Chu-Song Chen}, {and} \bibinfo{person}{Jie Zhou}.}
  \bibinfo{year}{2016}\natexlab{}.
\newblock \showarticletitle{Learning compact binary descriptors with
  unsupervised deep neural networks}. In \bibinfo{booktitle}{\emph{Proceedings
  of the IEEE Conference on Computer Vision and Pattern Recognition}}.
  \bibinfo{pages}{1183--1192}.
\newblock


\bibitem[\protect\citeauthoryear{Lin, Yang, Wang, and Piramuthu}{Lin
  et~al\mbox{.}}{2018}]%
        {lin2018adversarial}
\bibfield{author}{\bibinfo{person}{Kevin Lin}, \bibinfo{person}{Fan Yang},
  \bibinfo{person}{Qiaosong Wang}, {and} \bibinfo{person}{Robinson Piramuthu}.}
  \bibinfo{year}{2018}\natexlab{}.
\newblock \showarticletitle{Adversarial Learning for Fine-grained Image
  Search}.
\newblock \bibinfo{journal}{\emph{arXiv preprint arXiv:1807.02247}}
  (\bibinfo{year}{2018}).
\newblock


\bibitem[\protect\citeauthoryear{Lin, Doll{\'a}r, Girshick, He, Hariharan, and
  Belongie}{Lin et~al\mbox{.}}{2017}]%
        {lin2017feature}
\bibfield{author}{\bibinfo{person}{Tsung-Yi Lin}, \bibinfo{person}{Piotr
  Doll{\'a}r}, \bibinfo{person}{Ross~B Girshick}, \bibinfo{person}{Kaiming He},
  \bibinfo{person}{Bharath Hariharan}, {and} \bibinfo{person}{Serge~J
  Belongie}.} \bibinfo{year}{2017}\natexlab{}.
\newblock \showarticletitle{Feature Pyramid Networks for Object Detection.}. In
  \bibinfo{booktitle}{\emph{Proceedings of the IEEE conference on computer
  vision and pattern recognition}}, Vol.~\bibinfo{volume}{1}.
  \bibinfo{pages}{4}.
\newblock


\bibitem[\protect\citeauthoryear{Liu, Wang, Shan, and Chen}{Liu
  et~al\mbox{.}}{2016}]%
        {liu2016deep}
\bibfield{author}{\bibinfo{person}{Haomiao Liu}, \bibinfo{person}{Ruiping
  Wang}, \bibinfo{person}{Shiguang Shan}, {and} \bibinfo{person}{Xilin Chen}.}
  \bibinfo{year}{2016}\natexlab{}.
\newblock \showarticletitle{Deep supervised hashing for fast image retrieval}.
  In \bibinfo{booktitle}{\emph{Proceedings of the IEEE conference on computer
  vision and pattern recognition}}. \bibinfo{pages}{2064--2072}.
\newblock


\bibitem[\protect\citeauthoryear{Liu, Wang, Ji, Jiang, and Chang}{Liu
  et~al\mbox{.}}{2012}]%
        {liu2012supervised}
\bibfield{author}{\bibinfo{person}{Wei Liu}, \bibinfo{person}{Jun Wang},
  \bibinfo{person}{Rongrong Ji}, \bibinfo{person}{Yu-Gang Jiang}, {and}
  \bibinfo{person}{Shih-Fu Chang}.} \bibinfo{year}{2012}\natexlab{}.
\newblock \showarticletitle{Supervised hashing with kernels}. In
  \bibinfo{booktitle}{\emph{Proceedings of the IEEE conference on computer
  vision and pattern recognition}}. IEEE, \bibinfo{pages}{2074--2081}.
\newblock


\bibitem[\protect\citeauthoryear{Liu, Wang, Kumar, and Chang}{Liu
  et~al\mbox{.}}{2011}]%
        {liu2011hashing}
\bibfield{author}{\bibinfo{person}{Wei Liu}, \bibinfo{person}{Jun Wang},
  \bibinfo{person}{Sanjiv Kumar}, {and} \bibinfo{person}{Shih-Fu Chang}.}
  \bibinfo{year}{2011}\natexlab{}.
\newblock \showarticletitle{Hashing with graphs}. In
  \bibinfo{booktitle}{\emph{Proceedings of the 28th international conference on
  machine learning (ICML-11)}}. \bibinfo{pages}{1--8}.
\newblock


\bibitem[\protect\citeauthoryear{Norouzi and Blei}{Norouzi and Blei}{2011}]%
        {norouzi2011minimal}
\bibfield{author}{\bibinfo{person}{Mohammad Norouzi} {and}
  \bibinfo{person}{David~M Blei}.} \bibinfo{year}{2011}\natexlab{}.
\newblock \showarticletitle{Minimal loss hashing for compact binary codes}. In
  \bibinfo{booktitle}{\emph{Proceedings of the 28th international conference on
  machine learning (ICML-11)}}. Citeseer, \bibinfo{pages}{353--360}.
\newblock


\bibitem[\protect\citeauthoryear{Paszke, Gross, Chintala, and Chanan}{Paszke
  et~al\mbox{.}}{2017}]%
        {paszke2017pytorch}
\bibfield{author}{\bibinfo{person}{Adam Paszke}, \bibinfo{person}{Sam Gross},
  \bibinfo{person}{Soumith Chintala}, {and} \bibinfo{person}{Gregory Chanan}.}
  \bibinfo{year}{2017}\natexlab{}.
\newblock \bibinfo{title}{PyTorch}.
\newblock
\newblock


\bibitem[\protect\citeauthoryear{Qiu, Pan, Yao, and Mei}{Qiu
  et~al\mbox{.}}{2017}]%
        {qiu2017deep}
\bibfield{author}{\bibinfo{person}{Zhaofan Qiu}, \bibinfo{person}{Yingwei Pan},
  \bibinfo{person}{Ting Yao}, {and} \bibinfo{person}{Tao Mei}.}
  \bibinfo{year}{2017}\natexlab{}.
\newblock \showarticletitle{Deep semantic hashing with generative adversarial
  networks}. In \bibinfo{booktitle}{\emph{Proceedings of the 40th International
  ACM SIGIR Conference on Research and Development in Information Retrieval}}.
  ACM, \bibinfo{pages}{225--234}.
\newblock


\bibitem[\protect\citeauthoryear{Ronneberger, Fischer, and Brox}{Ronneberger
  et~al\mbox{.}}{2015}]%
        {ronneberger2015u}
\bibfield{author}{\bibinfo{person}{Olaf Ronneberger}, \bibinfo{person}{Philipp
  Fischer}, {and} \bibinfo{person}{Thomas Brox}.}
  \bibinfo{year}{2015}\natexlab{}.
\newblock \showarticletitle{U-net: Convolutional networks for biomedical image
  segmentation}. In \bibinfo{booktitle}{\emph{International Conference on
  Medical image computing and computer-assisted intervention}}. Springer,
  \bibinfo{pages}{234--241}.
\newblock


\bibitem[\protect\citeauthoryear{Salakhutdinov and Hinton}{Salakhutdinov and
  Hinton}{2007}]%
        {salakhutdinov2007learning}
\bibfield{author}{\bibinfo{person}{Ruslan Salakhutdinov} {and}
  \bibinfo{person}{Geoff Hinton}.} \bibinfo{year}{2007}\natexlab{}.
\newblock \showarticletitle{Learning a nonlinear embedding by preserving class
  neighbourhood structure}. In \bibinfo{booktitle}{\emph{Artificial
  Intelligence and Statistics}}. \bibinfo{pages}{412--419}.
\newblock


\bibitem[\protect\citeauthoryear{Shen, Xu, Liu, Yang, Huang, and Shen}{Shen
  et~al\mbox{.}}{2018}]%
        {shen2018unsupervised}
\bibfield{author}{\bibinfo{person}{Fumin Shen}, \bibinfo{person}{Yan Xu},
  \bibinfo{person}{Li Liu}, \bibinfo{person}{Yang Yang}, \bibinfo{person}{Zi
  Huang}, {and} \bibinfo{person}{Heng~Tao Shen}.}
  \bibinfo{year}{2018}\natexlab{}.
\newblock \showarticletitle{Unsupervised deep hashing with similarity-adaptive
  and discrete optimization}.
\newblock \bibinfo{journal}{\emph{IEEE transactions on pattern analysis and
  machine intelligence}} (\bibinfo{year}{2018}).
\newblock


\bibitem[\protect\citeauthoryear{Simonyan and Zisserman}{Simonyan and
  Zisserman}{2014}]%
        {simonyan2014very}
\bibfield{author}{\bibinfo{person}{Karen Simonyan} {and}
  \bibinfo{person}{Andrew Zisserman}.} \bibinfo{year}{2014}\natexlab{}.
\newblock \showarticletitle{Very deep convolutional networks for large-scale
  image recognition}.
\newblock \bibinfo{journal}{\emph{arXiv preprint arXiv:1409.1556}}
  (\bibinfo{year}{2014}).
\newblock


\bibitem[\protect\citeauthoryear{Wang, Kumar, and Chang}{Wang
  et~al\mbox{.}}{2010a}]%
        {wang2010semi}
\bibfield{author}{\bibinfo{person}{Jun Wang}, \bibinfo{person}{Sanjiv Kumar},
  {and} \bibinfo{person}{Shih-Fu Chang}.} \bibinfo{year}{2010}\natexlab{a}.
\newblock \showarticletitle{Semi-supervised hashing for scalable image
  retrieval}.
\newblock  (\bibinfo{year}{2010}).
\newblock


\bibitem[\protect\citeauthoryear{Wang, Kumar, and Chang}{Wang
  et~al\mbox{.}}{2010b}]%
        {wang2010sequential}
\bibfield{author}{\bibinfo{person}{Jun Wang}, \bibinfo{person}{Sanjiv Kumar},
  {and} \bibinfo{person}{Shih-Fu Chang}.} \bibinfo{year}{2010}\natexlab{b}.
\newblock \showarticletitle{Sequential projection learning for hashing with
  compact codes}. In \bibinfo{booktitle}{\emph{Proceedings of the 27th
  international conference on machine learning (ICML-10)}}.
  \bibinfo{pages}{1127--1134}.
\newblock


\bibitem[\protect\citeauthoryear{Wang, Zhang, Sebe, Shen, et~al\mbox{.}}{Wang
  et~al\mbox{.}}{2018}]%
        {wang2018survey}
\bibfield{author}{\bibinfo{person}{Jingdong Wang}, \bibinfo{person}{Ting
  Zhang}, \bibinfo{person}{Nicu Sebe}, \bibinfo{person}{Heng~Tao Shen},
  {et~al\mbox{.}}} \bibinfo{year}{2018}\natexlab{}.
\newblock \showarticletitle{A survey on learning to hash}.
\newblock \bibinfo{journal}{\emph{IEEE Transactions on Pattern Analysis and
  Machine Intelligence}} \bibinfo{volume}{40}, \bibinfo{number}{4}
  (\bibinfo{year}{2018}), \bibinfo{pages}{769--790}.
\newblock


\bibitem[\protect\citeauthoryear{Wang, Zhang, and Si}{Wang
  et~al\mbox{.}}{2015}]%
        {wang2015ranking}
\bibfield{author}{\bibinfo{person}{Qifan Wang}, \bibinfo{person}{Zhiwei Zhang},
  {and} \bibinfo{person}{Luo Si}.} \bibinfo{year}{2015}\natexlab{}.
\newblock \showarticletitle{Ranking Preserving Hashing for Fast Similarity
  Search.}. In \bibinfo{booktitle}{\emph{Proceedings of the International Joint
  Conference on Artificial Intelligence}}. \bibinfo{pages}{3911--3917}.
\newblock


\bibitem[\protect\citeauthoryear{Weiss, Torralba, and Fergus}{Weiss
  et~al\mbox{.}}{2009}]%
        {weiss2009spectral}
\bibfield{author}{\bibinfo{person}{Yair Weiss}, \bibinfo{person}{Antonio
  Torralba}, {and} \bibinfo{person}{Rob Fergus}.}
  \bibinfo{year}{2009}\natexlab{}.
\newblock \showarticletitle{Spectral hashing}. In
  \bibinfo{booktitle}{\emph{Advances in neural information processing
  systems}}. \bibinfo{pages}{1753--1760}.
\newblock


\bibitem[\protect\citeauthoryear{Welinder, Branson, Mita, Wah, Schroff,
  Belongie, and Perona}{Welinder et~al\mbox{.}}{2010}]%
        {welinder2010caltech}
\bibfield{author}{\bibinfo{person}{Peter Welinder}, \bibinfo{person}{Steve
  Branson}, \bibinfo{person}{Takeshi Mita}, \bibinfo{person}{Catherine Wah},
  \bibinfo{person}{Florian Schroff}, \bibinfo{person}{Serge Belongie}, {and}
  \bibinfo{person}{Pietro Perona}.} \bibinfo{year}{2010}\natexlab{}.
\newblock \showarticletitle{Caltech-UCSD birds 200}.
\newblock  (\bibinfo{year}{2010}).
\newblock


\bibitem[\protect\citeauthoryear{Wu, Zhu, Cai, Chen, and Bu}{Wu
  et~al\mbox{.}}{2013}]%
        {wu2013semi}
\bibfield{author}{\bibinfo{person}{Chenxia Wu}, \bibinfo{person}{Jianke Zhu},
  \bibinfo{person}{Deng Cai}, \bibinfo{person}{Chun Chen}, {and}
  \bibinfo{person}{Jiajun Bu}.} \bibinfo{year}{2013}\natexlab{}.
\newblock \showarticletitle{Semi-supervised nonlinear hashing using bootstrap
  sequential projection learning}.
\newblock \bibinfo{journal}{\emph{IEEE Transactions on Knowledge and Data
  Engineering}} \bibinfo{volume}{25}, \bibinfo{number}{6}
  (\bibinfo{year}{2013}), \bibinfo{pages}{1380--1393}.
\newblock


\bibitem[\protect\citeauthoryear{Xia, Pan, Lai, Liu, and Yan}{Xia
  et~al\mbox{.}}{2014}]%
        {xia2014supervised}
\bibfield{author}{\bibinfo{person}{Rongkai Xia}, \bibinfo{person}{Yan Pan},
  \bibinfo{person}{Hanjiang Lai}, \bibinfo{person}{Cong Liu}, {and}
  \bibinfo{person}{Shuicheng Yan}.} \bibinfo{year}{2014}\natexlab{}.
\newblock \showarticletitle{Supervised Hashing for Image Retrieval via Image
  Representation Learning.}. In \bibinfo{booktitle}{\emph{Proceedings of the
  AAAI Conference on Artificial Intelligence}}, Vol.~\bibinfo{volume}{1}.
  \bibinfo{pages}{2156--2162}.
\newblock


\bibitem[\protect\citeauthoryear{Yu, Sun, Yang, Rui, and Yao}{Yu
  et~al\mbox{.}}{2018}]%
        {yu2018hierarchical}
\bibfield{author}{\bibinfo{person}{Wei Yu}, \bibinfo{person}{Xiaoshuai Sun},
  \bibinfo{person}{Kuiyuan Yang}, \bibinfo{person}{Yong Rui}, {and}
  \bibinfo{person}{Hongxun Yao}.} \bibinfo{year}{2018}\natexlab{}.
\newblock \showarticletitle{Hierarchical semantic image matching using CNN
  feature pyramid}.
\newblock \bibinfo{journal}{\emph{Computer Vision and Image Understanding}}
  \bibinfo{volume}{169} (\bibinfo{year}{2018}), \bibinfo{pages}{40--51}.
\newblock


\bibitem[\protect\citeauthoryear{Yunchao, Lazebnik, Gordo, and
  Perronnin}{Yunchao et~al\mbox{.}}{2013}]%
        {yunchao2013iterative}
\bibfield{author}{\bibinfo{person}{Gong Yunchao}, \bibinfo{person}{S Lazebnik},
  \bibinfo{person}{A Gordo}, {and} \bibinfo{person}{F Perronnin}.}
  \bibinfo{year}{2013}\natexlab{}.
\newblock \showarticletitle{Iterative quantization: a procrustean approach to
  learning binary codes for large-scale image retrieval}.
\newblock \bibinfo{journal}{\emph{IEEE transactions on pattern analysis and
  machine intelligence}} \bibinfo{volume}{35}, \bibinfo{number}{12}
  (\bibinfo{year}{2013}), \bibinfo{pages}{2916--2929}.
\newblock


\bibitem[\protect\citeauthoryear{Zhang, Lai, and Feng}{Zhang
  et~al\mbox{.}}{2018}]%
        {zhang2018attention}
\bibfield{author}{\bibinfo{person}{Xi Zhang}, \bibinfo{person}{Hanjiang Lai},
  {and} \bibinfo{person}{Jiashi Feng}.} \bibinfo{year}{2018}\natexlab{}.
\newblock \showarticletitle{Attention-Aware Deep Adversarial Hashing for
  Cross-Modal Retrieval}. In \bibinfo{booktitle}{\emph{Proceedings of the
  European Conference on Computer Vision}}. \bibinfo{pages}{614--629}.
\newblock


\bibitem[\protect\citeauthoryear{Zhao, Luo, Peng, and Fan}{Zhao
  et~al\mbox{.}}{2017}]%
        {zhao2017spatial}
\bibfield{author}{\bibinfo{person}{Wanqing Zhao}, \bibinfo{person}{Hangzai
  Luo}, \bibinfo{person}{Jinye Peng}, {and} \bibinfo{person}{Jianping Fan}.}
  \bibinfo{year}{2017}\natexlab{}.
\newblock \showarticletitle{Spatial pyramid deep hashing for large-scale image
  retrieval}.
\newblock \bibinfo{journal}{\emph{Neurocomputing}}  \bibinfo{volume}{243}
  (\bibinfo{year}{2017}), \bibinfo{pages}{166--173}.
\newblock


\bibitem[\protect\citeauthoryear{Zhuang, Lin, Shen, and Reid}{Zhuang
  et~al\mbox{.}}{2016}]%
        {zhuang2016fast}
\bibfield{author}{\bibinfo{person}{Bohan Zhuang}, \bibinfo{person}{Guosheng
  Lin}, \bibinfo{person}{Chunhua Shen}, {and} \bibinfo{person}{Ian Reid}.}
  \bibinfo{year}{2016}\natexlab{}.
\newblock \showarticletitle{Fast training of triplet-based deep binary
  embedding networks}. In \bibinfo{booktitle}{\emph{Proceedings of the IEEE
  Conference on Computer Vision and Pattern Recognition}}.
  \bibinfo{pages}{5955--5964}.
\newblock


\end{thebibliography}

\end{document}